\documentclass{article} % For LaTeX2e
\pdfoutput=1
\usepackage{natbib}
\usepackage{amsmath}
\usepackage{amssymb}
\usepackage{times}
\usepackage{graphicx}
\usepackage{subfigure}

\newcommand{\fracpartial}[2]{\frac{\partial #1}{\partial  #2}}
\newcommand{\fracpartialdouble}[2]{\frac{\partial^2 #1}{\partial  #2^2}}
\newcommand{\fracpartialij}[2]{\frac{\partial^2 #1}{\partial #2^{[i]}\partial #2^{[j]}}}

\newcommand{\Lkl}{\mathcal{L}_{\text{KL}}}
\newcommand{\Lmf}{\mathcal{L}_{\text{MF}}}

\newcommand{\amo}{\alpha^o_m}
\newcommand{\hao}{\hat \alpha^o}
\newcommand{\bR}{\mathbf R}
\newcommand{\bT}{\mathbf T}
\newcommand{\sumover}[1]{\sum_{#1=1}^{\MakeUppercase #1}}
\newcommand{\bell}{\boldsymbol \ell}
\newcommand{\bL}{  {\bf L} }

\newcommand{\bZ}{  {\bf Z} }
\newcommand{\bS}{  {\bf S} }
\newcommand{\bs}{  {\bf s} }

\newcommand{\bX}{  {\bf X} }

\newcommand{\bY}{  {\bf Y} }
\newcommand{\by}{  {\bf y} }

\newcommand{\bg}{  {\bf g} }
\newcommand{\bm}{  {\bf m} }

\newcommand{\bEta}{  {\boldsymbol \eta} }
\newcommand{\bmu}{  {\boldsymbol \mu} }
\newcommand{\bpi}{  {\boldsymbol \pi} }
\newcommand{\br}{  {\bf r} }

\newcommand{\balpha}{  {\boldsymbol \alpha} }
\newcommand{\btheta}{  {\boldsymbol \theta} }
\newcommand{\bLambda}{  {\boldsymbol \Lambda} }
\newcommand{\rhat}{\widehat r}

\newcommand{\ybar}{\bar{\bf y}}
\renewcommand{\digamma}{\psi}
\newcommand{\<}{\langle}
\renewcommand{\>}{\rangle}
\renewcommand{\d}{\,\text d}
\newcommand{\Dir}{\textit{Dir}}
\newcommand{\given}{\,|\,}

\newcommand{\prd}[1]{\prod_{#1=1}^{\MakeUppercase #1}}

\newcommand{\E}[2]{\mathbb E_{#1}\Big [{#2}\Big ]}

\title{Fast Variational Inference in the Conjugate Exponential Family}

\author{ James Hensman, Magnus Rattray and Neil D. Lawrence}
\begin{document}

\maketitle

\begin{abstract}
  We present a general method for deriving collapsed
  variational inference algorithms for probabilistic models in the
  conjugate exponential family. Our method unifies many existing
  approaches to collapsed variational inference. Our collapsed
  variational inference leads to a new lower bound on the marginal
  likelihood. % We show that standard variational expectation
  % maximization performs \emph{steepest ascent} on the new bound.
  We exploit the information geometry of the bound to derive much
  faster optimization methods based on conjugate gradients for these
  models. Our approach is very general and is easily applied to any model
  where the mean field update equations have been
  derived. Empirically we show significant speed-ups for
  probabilistic inference using our bound.
\end{abstract}

\section{Introduction}
\label{introduction}

Variational bounds provide a convenient approach to approximate
inference in a range of intractable models. Classical variational
optimisation is achieved through coordinate ascent
which can be slow to converge. A popular solution
\citep{king2006fast,teh2007collapsed,kurihara2007collapsed,sung2008latent,lazaro2011variational,lazaro2011overlapping}
is to marginalize analytically a portion of the variational
approximating distribution, removing this from the optimization. In
this paper we provide a unifying framework for collapsed inference in
the general class of models composed of conjugate-exponential graphs
(CEGs).
 %The common strand to all these approaches is that a subset of variables
%are analytically marginalised at some point when constructing the
%variational bound.

First we review the body of earlier work with a succinct and unifying
derivation of the collapsed bounds. We describe how the applicability of
the collapsed bound to any particular CEG can be determined with a
simple d-separation test. Standard variational inference via
\emph{coordinate ascent} turns out to be \emph{steepest ascent} with a
unit step length on our unifying bound. This motivates us to consider
natural gradients and conjugate gradients for fast optimization of
these models. We apply our unifying approach to a range of models from
the literature obtaining, often, an order of magnitude or more
increase in convergence speed. Our unifying view allows collapsed
variational methods to be integrated into general inference tools like
infer.net \citep{minka2010infer}.

 %If the model
%exhibits strong correlations between parameters (which is often the
%case) the deficiencies of both coordinate ascent and steepest ascent
%for models are well understood. %The popularity of the range of
%different collapsed bounds we unify stems from the fact that we can
%now consider much more advanced optimization algorithms. We show
%empirically that these optimization algorithms converge much quicker
%with an application of the framework to a Bayesian mixture
%model. First though we re-derive the broad class of
%collapsed/marginalised variational bounds from the KL corrected
%perspective.

\section{The Marginalised Variational Bound}
\label{sec:mvb}

The advantages to marginalising analytically a subset of variables in
variational bounds seem to be well understood: several different
approaches have been suggested in the context of specific models. In
Dirichlet process mixture models \citet{kurihara2007collapsed}
proposed a collapsed approach using both truncated stick-breaking and
symmetric priors. \citet{sung2008latent} proposed `latent space
variational Bayes' where both the cluster-parameters and mixing
weights were marginalised, again with some
approximations. \citet{teh2007collapsed} proposed a collapsed
inference procedure for latent Dirichlet allocation (LDA).
% , where the posterior for the latent variables $q(\bZ)$
           % was assumed independent of the posterior over the topic
           % models and document models ($\phi, \theta$).
% Teh showed that free-form inference of the topic and document models
% % $\phi, \theta$
% was equivalent to marginalising them.
In this paper we unify all these results from the perspective of the
`KL corrected bound' \citep{king2006fast}. This lower bound on the
model evidence is also an upper bound on the original variational
bound, the difference between the two bounds is given by a Kullback
Leibler divergence. The approach has also been referred to as the
\emph{marginalised variational bound} by
\citet{lazaro2011overlapping,lazaro2011variational}. % Collapsed
% variational approaches are also very similar in spirit: they are
% referred to as collapsed by analogy with collapsed Gibbs samplers,
% which also analytically marginalise a subset of the variables.
% Collapsed variational methods have been proposed for latent variable
% models \citep{sung2008latent} mixtures \citep{kurihara2007collapsed}
% and topic models \citep{teh2007collapsed}.; The new derivation is
% succinct and general. 
The connection between the KL corrected bound and the collapsed
bounds is not immediately obvious. The key difference between the
frameworks is the order in which the marginalisation and variational
approximation are applied. However, for CEGs this order turns out to
be irrelevant. Our framework leads to a more succinct
derivation of the collapsed approximations. The resulting bound can
then be optimised without recourse to approximations in either the
bound's evaluation or its optimization.

%\james{:is it worth spelling out our contributions? New deirvation. Unify with CVB. Equivalence of gradient. Curvature. Riemannian applied to colapsed problem. Riemannian trick to avoid FI inverse. }

% In section \ref{sec:riemann}, we use approximate Riemannian optimisation \citep{honkela2010approximate} to improve the speed of inference. The combination of the MV bound and natural gradients is especially potent, since the MV bound reduces the number of variational parameters. We also derive an expression for the natural gradient of factorising multinomial distributions which does not require a matrix inverse, contrary to previous implementations.  We thus provide an optimisation procedure based on conjugate natural-gradient ascent which significantly reduces computational time. 

\subsection{Variational Inference}

% Variational Bayes (VB)  is a
% method for approximate Bayesian inference where the posterior
% distribution of the parameters is approximated by a distribution which
% renders the required Bayesian integrals tractable.
Assume we have a probabilistic model for data, $\mathcal D$, given
parameters (and/or latent variables), $\bX$, $\bZ$, of the form
$p(\mathcal D, \bX,\bZ) = p(\mathcal D \,|\, \bZ, \bX) p(\bZ\,|\,\bX)p(\bX)$.  In
variational Bayes (see e.g. \citet{bishop2006pattern}) we approximate
the posterior $p(\bZ,\bX| \mathcal D)$ by a distribution
$q(\bZ,\bX)$. We use Jensen's inequality to derive a lower bound on
the model evidence $\mathcal L$, which serves as an objective function
in the variational optimisation:
\begin{equation}
  p(\mathcal D) \geq \mathcal L = \int \!q(\bZ,\bX) \ln \frac{p(\mathcal D, \bZ, \bX)}{q(\bZ,\bX)} \text{d}\bZ\,\text{d}\bX.
  \label{eq:lower_bound}
\end{equation}
For tractability the mean field (MF) approach assumes $q$ factorises
across its variables, $q(\bZ, \bX) = q(\bZ)q(\bX)$.  It is then
possible to implement an optimisation scheme which analytically
optimises each factor alternately, %\citep{bishop2006pattern}
with the optimal distribution given by
\begin{equation}
	q^\star(\bX) \propto \exp\left\{\int \! q(\bZ)\ln p(\mathcal D,\bX| \bZ)\,\text{d}\bZ\right\}, 
	\label{eq:updates}
\end{equation}
and similarly for $\bZ$: these are often referred to as VBE and VBM
steps. \citet{king2006fast} substituted the expression for the optimal
distribution (for example $q^\star(\bX)$) back into the bound
\eqref{eq:lower_bound}, eliminating one set of parameters from the
optimisation, an approach that has been reused by
\citet{lazaro2011overlapping,lazaro2011variational}. The resulting
bound is not dependent on $q(\bX)$.  \citet{king2006fast} referred to
this new bound as `the KL corrected bound'. The difference between the
bound, which we denote $\Lkl$, and a standard mean field
approximation $\Lmf$, is the Kullback Leibler divergence between the
optimal form of $q^*(\bX)$ and the current $q(\bX)$.%approximation, $q(\bX)$.
 
We rederive their bound % This derivation first made a fully mean-field approximation
% $q(\bZ)q(\bX)$ and then manipulated the expression along with the
% relation \eqref{eq:updates} to remove one of the factors of this
% variational posterior.  A more general derivation can be constructed
by first
using Jensen's inequality to construct the variational lower
bound on the {\em conditional} distribution, 
\begin{equation}
  \ln p(\mathcal D|\bX) \geq \int\! q(\bZ)\ln \frac{p(\mathcal D,\bZ|\bX)}{q(\bZ)}\,\text{d}\bZ \triangleq \mathcal L_1 . 
\label{eq:L_1}
\end{equation}
This object turns out to be of central importance in computing the final KL-corrected bound and also in computing gradients, curvatures and the distribution of the collapsed variables $q^\star(\bX)$. It is easy to see that it is a function of $\bX$ which lower-bounds the log likelihood $p(\mathcal D\given\bX)$, and indeed our derivation treats it as such. We now
marginalize the conditioned variable from this expression, % First the
% conditional bound: Taking \eqref{eq:L_1} and exponentiating both sides, multiplying by $p(\bX)$,
% integrating and taking the logarithm (all operations for which the
% inequality holds) we find
\begin{equation}
  \ln p(\mathcal D) \geq \ln \int\! p(\bX)\exp \{\mathcal L_1\}\,\text{d}\bX \triangleq \Lkl,
	\label{eq:MVB_derivation}
\end{equation}
giving us the bound of \citet{king2006fast} \&
\citet{lazaro2011overlapping}. Note that one set of parameters was
marginalised \emph{after} the variational approximation was made.
%We stick with the term `Marginalised Variational Bound' since it is clear from the above that one set of parameters is marginalised, {\em after} the variational approximation is made. 

Using \eqref{eq:updates}, this expression also
provides the approximate posterior for the marginalised
variables $\bX$:
\begin{equation}
	q^\star(\bX) = p(\bX)e^{\mathcal L_1-\Lkl}
	\label{eq:qstar_relation}
\end{equation}
and $e^{\mathcal \Lkl}$ appears as the constant of proportionality in the mean-field update equation \eqref{eq:updates}.

%\begin{equation}
  %\int\! p(\bX)\exp \{\mathcal L_1\}\d\bX = \int\!\exp\{\mathcal L_{KL}\} q^\star(\bX) \d \bX
	%\label{eq:MVB_missing_dist}
%\end{equation}
%thus the constant of proportionality in the update rule \eqref{eq:updates} is the exponent of $\mathcal L_{KL}$.  
%

\section{Partial Equivalence of the Bounds}

We can recover $\Lmf$ from $\Lkl$ by again applying
Jensen's inequality,% .  Taking equation \eqref{eq:MVB_derivation} and
% introducing $q(\bX)$, we can apply Jensen's inequality:
\begin{equation}
	\Lkl = \ln\int\! q(\bX)\frac{p(\bX)}{q(\bX)}\exp\{\mathcal L_1\}\,\text{d}\bX \geq
	\int\! q(\bX)\ln\left\{\frac{p(\bX)}{q(\bX)}\exp\{\mathcal L_1\}\right\}\,\text{d}\bX,
\end{equation}
%Noting that this is the log of an expectation under $q(\bX)$, we apply Jensen's inequality and obtain
%\begin{equation}
	%\mathcal L_{KL} \geq \int\! q(\bX)\ln\left\{\frac{p(\bX)}{q(\bX)}\exp\{\mathcal L_1\}\right\}\,\text{d}\bX,
%\end{equation}
which can be re-arranged to give the mean-field bound, 
\begin{equation}
	\Lkl \geq \int\! q(\bX)q(\bZ)\ln\left\{\frac{p(\mathcal D|\bZ,\bX)p(\bZ)p(\bX)}{q(\bZ)q(\bX)}\right\}\,\text{d}\bX\,\text{d}\bZ,
\end{equation}
and it follows that $\Lkl = \Lmf+\text{KL}(q^*(\bX)||q(\bX))$ and\footnote{We use $\text{KL}(\cdot||\cdot)$ to denote the Kullback Leibler divergence between two distributions.} $\Lkl \geq
\Lmf$.  For a given $q(\bZ)$, the bounds are equal after
$q(\bX)$ is updated via the mean field method: the approximations are ultimately the same. 
The advantage of the new bound is to reduce the number of
parameters in the optimisation. It is particularly
useful when variational parameters are optimised by gradient
methods. Since VBEM is equivalent to a steepest descent gradient method with a fixed step
size, there appears to be a lot to gain by combining the KLC bound
with more sophisticated optimization techniques.

\subsection{Gradients}
Consider the gradient of the KL corrected bound with respect to the parameters of $q(\bZ)$:
\begin{equation}
\fracpartial \Lkl {\theta_z} = \exp\{-\Lkl\}\fracpartial {} {\theta_z }\int\! \exp\{\mathcal L_1\}p(\bX) \d \bX = \E {q^\star(\bX)} {\fracpartial {\mathcal L_1} {\theta_z} },
\end{equation}
where we have used the relation \eqref{eq:qstar_relation}.  To find
the gradient of the mean-field bound we note that it can be written in
terms of our conditional bound \eqref{eq:L_1} as $\Lmf = \E {q(\bX)}
{\mathcal L_1 + \ln p(\bX) - \ln q(\bX)}$ giving 
\begin{equation}
\fracpartial \Lmf {\theta_z} = \E {q(\bX)} {\fracpartial {\mathcal L_1} {\theta_z} }
\label{eq:MVB_grad}
\end{equation}
thus setting $q(\bX) = q^\star(\bX)$ not only makes the bounds equal, $\Lmf=\Lkl$, but also their \emph{gradients} with respect to $\theta_Z$.

\citet{sato2001online} has shown that the variational update equation
can be interpreted as a \emph{gradient} method, where each update is
also a step in the steepest direction in the canonical parameters of
$q(\bZ)$.  We can combine this important insight with the above result
to realize that we have a simple method for computing the gradients of
the KL corrected bound: we only need to look at the update
expressions for the mean-field method. This result also reveals the
weakness of standard variational Bayesian expectation maximization
(VBEM): it is a steepest ascent
algorithm. \citet{honkela2010approximate} looked to rectify this
weakness by applying a conjugate gradient algorithm to the mean field
bound. However, they didn't obtain a significant improvement in
convergence speed. Our suggestion is to apply conjugate gradients to
the KLC bound. Whilst the value and gradient of the MF bound matches
that of the KLC bound after an update of the collapsed variables, the
\emph{curvature} is always greater. In practise this means that much larger
steps (which we compute using conjugate gradient methods) can be taken when optimizing the KLC bound than for the MF
bound leading to more rapid convergence.

\subsection{Curvature of the Bounds}

\citet{king2006fast} showed empirically that the KLC bound could lead
to faster convergence because the bounds differ in their curvature:
the curvature of the KLC bound enables larger steps to be taken by an
optimizer. We now derive analytical expressions for the curvature of
both bounds. For the mean field bound we have%\neil{The mean field bound is still univariate!}
\begin{equation}
\fracpartialdouble \Lmf {\theta_z} = \E {q(\bX)} {\fracpartialdouble {\mathcal L_1} {\theta_z} },
\label{eq:MF_curvature}
\end{equation}
and for the KLC bound, with some manipulation of
\eqref{eq:MVB_derivation} and using \eqref{eq:qstar_relation}: %\neil{Shouldn't we have this curvature for the multivariate case?}
\begin{equation}
\begin{split}
\fracpartialij \Lkl {\theta_z} &= e^{-\Lkl}\fracpartialij {e^{\Lkl}} {\theta_z} - e^{-2\Lkl}\Big\{ \fracpartial{e^{\Lkl}}{\theta_z^{[i]}}\Big\}\Big\{ \fracpartial{e^{\Lkl}}{\theta_z^{[j]}}\Big\}\\
%&= e^{-\Lkl}\int\! p(\bX)e^{\mathcal L_1}\Big\{ \fracpartialij{{\mathcal L_1}}{\theta_z} + \fracpartial {\mathcal L_1} {\theta_z^{[i]}}\fracpartial {\mathcal L_1} {\theta_z^{[j]}}\Big\}\d \bX  - e^{-2\Lkl}\int\!p(\bX)\fracpartial{e^{\mathcal L_1}}{\theta_z}\d \bX\int\!p(\bX)\fracpartial{e^{\mathcal L_1}}{\theta_z}\d \bX \\
&= \E {q^\star(\bX)} {\fracpartialij {{\mathcal L_1}}{\theta_z}} +  \E {q^\star(\bX)} {\fracpartial {\mathcal L_1} {\theta_z^{[i]}}\fracpartial {\mathcal L_1} {\theta_z^{[j]}}} - \Big\{\E {q^\star(\bX)} {\fracpartial{\mathcal L_1}{\theta_z^{[i]}}} \Big\}\Big\{\E {q^\star(\bX)} {\fracpartial{\mathcal L_1}{\theta_z^{[j]}}} \Big\}.  
\end{split}
\label{eq:curvature}
\end{equation}
In this result the first term is equal to \eqref{eq:MF_curvature}, and
the second two terms combine to be always positive
semi-definite, % (via Jensen's inequality, this time considering expectations of a square)
proving \citet{king2006fast}'s intuition about the curvature of the
bound. When curvature is negative definite (e.g. near a maximum), the
KLC bound's curvature is less negative definite, enabling larger steps
to be taken in optimization. Figure \ref{fig:curvature} illustrates the effect of this as well as the bound's similarities.

\subsection{Relationship to Collapsed VB}

In \emph{collapsed inference} some parameters are marginalized {\em
  before} applying the variational bound. For example,
\citet{sung2008latent} proposed a latent variable model where the model
parameters were marginalised, and \citet{teh2007collapsed} proposed a
nonparametric topic model where the document proportions were collapsed. These procedures lead to improved inference, or faster convergence.

The KLC bound derivation we have provided also
marginalises parameters, but {\em after} a variational
approximation is made. The difference between the two approaches
is distilled in these expressions:\\
%\begin{eqnarray}
	%\ln \mathbb E_{p(\bX)}\bigg [ \exp \bigg\{ \mathbb E_{q(\bZ)} \big [\ln p(\mathcal D|\bX, \bZ) \big ]\bigg\}\bigg ]\label{mvb_part}\\
	%\mathbb E_{q(\bZ)}\bigg [ \ln \bigg\{ \mathbb E_{p(\bX)} \big [p(\mathcal D|\bX, \bZ) \big ]\bigg\}\bigg ]\label{cvb_part}
%\end{eqnarray}
%\begin{minipage}{0.5\linewidth}
\begin{equation}
	\ln \mathbb E_{p(\bX)}\bigg [ \exp \bigg\{ \mathbb E_{q(\bZ)} \big [\ln p(\mathcal D|\bX, \bZ) \big ]\bigg\}\bigg ]\,\,\,\,\,\,\,\,\,\,
%\end{minipage}
%\begin{minipage}{0.47\linewidth}
\mathbb E_{q(\bZ)}\bigg [ \ln \bigg\{ \mathbb E_{p(\bX)} \big [p(\mathcal D|\bX, \bZ) \big ]\bigg\}\bigg ]
\end{equation}
%\end{minipage}
%where \eqref{mvb_part} appears in the KLC bound, and \eqref{cvb_part}
where the left expression appears in the KLC bound, and the right expression
appears in the bound for collapsed variational Bayes, with the
remainder of the bounds being equal. Whilst appropriately conjugate
formulation of the model will always ensure that the KLC expression
%\eqref{mvb_part}
is analytically tractable, the expectation in the collapsed VB expression is not. 
%\eqref{cvb_part} is not. 
\citet{sung2008latent} propose a first order
approximation to the expectation of the form $ \E {q(\bZ)}{f(\bZ)}
\approx f(\E {q(\bZ)}{\bZ}), $ which reduces the right expression to the that on the left. 
Under this
approximation\footnote{\citet{kurihara2007collapsed} and
  \citet{teh2007collapsed} suggest a further second order
  correction and assume that that $q(\bZ)$ is Gaussian to obtain
  tractability. This leads to additional correction terms that augment
  KLC bound. The form of these corrections would need to be determined
  on a case by case basis, and has in fact been shown to be less effective than those methods unified here \citep{asuncion2012smoothing}.} % This approach was shown to work well
the KL corrected approach is equivalent to the collapsed variational
approach.
% for approximating discrete variables in mixture and topic models, but to our
% knowledge has not been applied more generally. \todo{Neil, need your persuasive
% skills here, potential point of attack for the reviewers}

%These two expressions can be
%shown to be equal if the log-likelihood is a linear combinations of factors of $\bX$ and $\bZ$, 
 %i.e. $\ln p(\mathcal D|\bX, \bZ) = \sum_i
%f_x^{(i)}(\mathcal D, \bX)f^{(i)}_z(\mathcal D, \bZ)$.  Clearly this
%is always the case for mixture models, since the likelihood term is
%$\ln p(\mathcal D|\bX,\bZ) = \sum_n\sum_k z_{nk} \ln p(\mathcal D_n|
%\bX_k)$, where $z_{nk}$ is the binary indicator assigning the
%$n^\text{th}$ datum to the $k^\text{th}$ cluster with parameters
%$\bX_k$.  Moreover, this equivalence holds for any model of the
%exponential form discussed by \citet{ghahramani2001propagation}.
%\magnus{I think the factors $f^i$ and $g^i$ also have to be
  %independent (I mean $\langle\prod_i g(f^{(i)}_x)\rangle = \prod_i
  %\langle g(f^{i}_x)\rangle_x$ and same for $f_z$) under the prior for
  %this to hold, which they are for a mixture model.}

\subsection{Applicability}

To apply the KLC bound we need to specify a subset, $\bX$, of
variables to marginalize. We select the variables that break the
dependency structure of the graph to enable the analytic computation
of the integral in \eqref{eq:MVB_derivation}. Assuming the appropriate
conjugate exponential structure for the model we are left with the
requirement to select a sub-set that induces the appropriate
factorisation. These induced factorisations are discussed in some
detail in \citet{bishop2006pattern}. They are factorisations in the
approximate posterior which arise from the form of the variational
approximation and from the structure of the model. These
factorisations allow application of KLC bound, and can be identified
using a simple d-separation test as Bishop discusses.

\begin{figure}
  \subfigure[\label{fig:dsep}]{\includegraphics[width=0.45\textwidth]{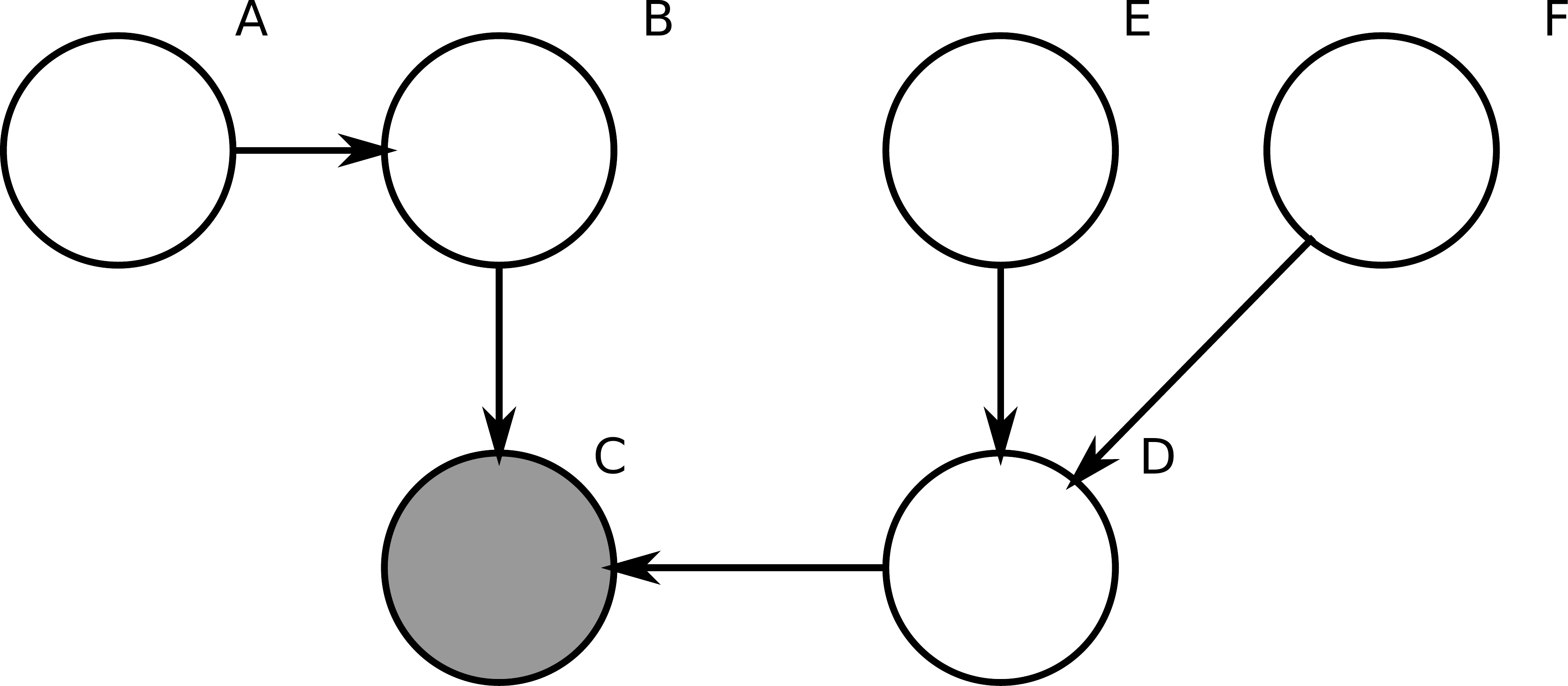}}\hfill\subfigure[\label{fig:curvature} ]{\includegraphics[width=0.45\textwidth]{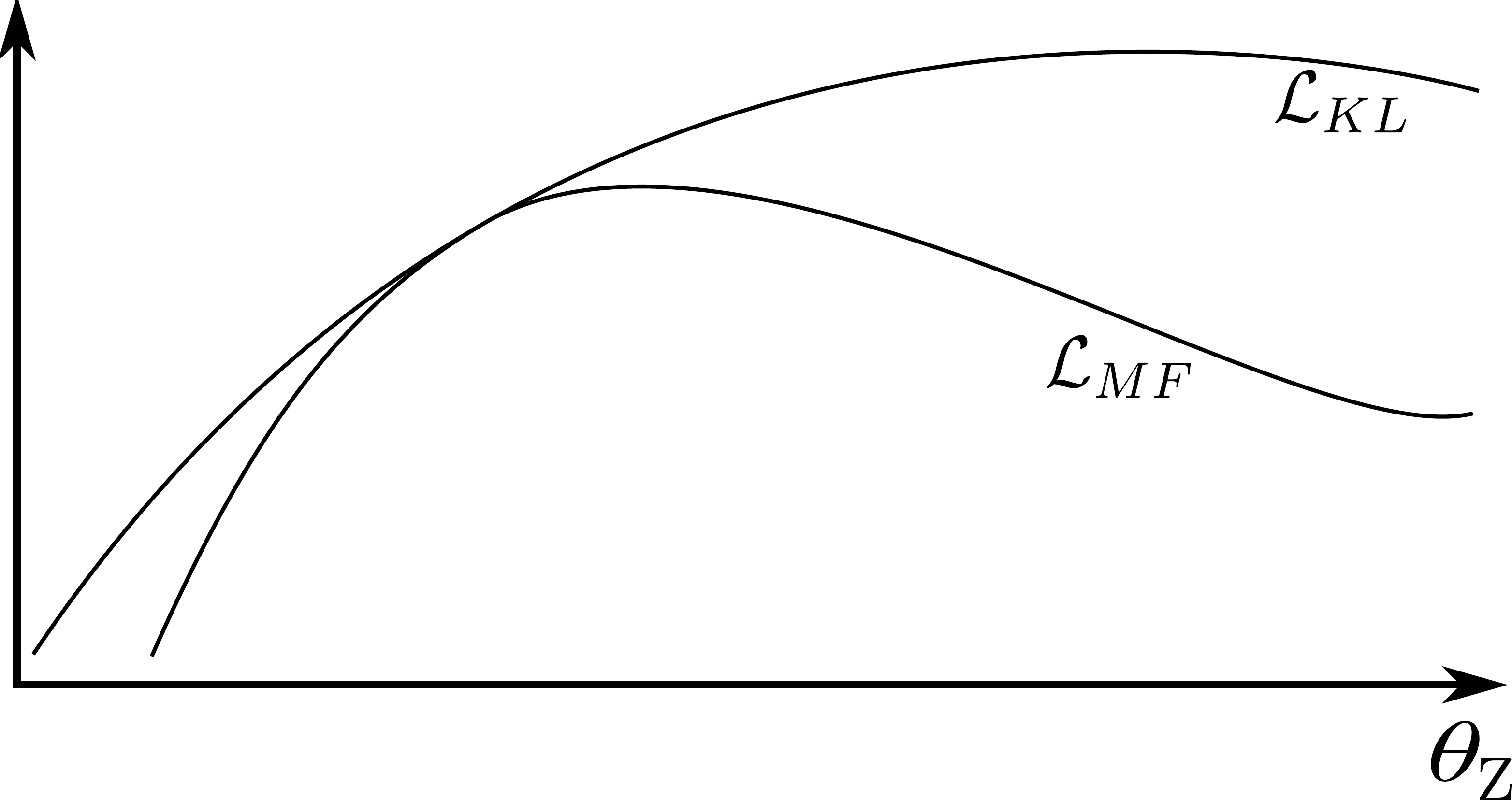}}
\caption{(a) An example directed graphical model on which we could use the KLC bound. Given the observed node C, the nodes A,F d-separate given nodes B,D,E. Thus we could make an explicit variational approximation for A,F, whilst marginalising B,D,E. Alternatively, we could select B,D,E for a parameterised approximate distribution, whilst marginalising A,F. (b) A sketch of the KLC and MF bounds. At the point where the mean field method has $q(\bX) = q^\star(\bX)$, the bounds are equal in value as well as in gradient. Away from the this point, the different between the bounds is the Kullback Leibler divergence between the current MF approximation for $\bX$ and the implicit distribution $q^\star(\bX)$ of the KLC bound.}
\end{figure}

The d-separation test involves checking for independence amongst the
marginalised variables ($\bX$ in the above) conditioned on the observed data
$\mathcal D$ and the approximated variables ($\bZ$ in the above). The requirement is to select a sufficient set of variables, $\bZ$, such that the effective
likelihood for $\bX$, given by \eqref{eq:L_1} becomes conjugate to the prior.  Figure \ref{fig:dsep} illustrates the d-separation test with application to the KLC bound.

For latent variable models, it is often sufficient to select the
latent variables for $\bX$ whilst collapsing the model variables. For
example, in the specific case of mixture models and topic models,
approximating the component labels allows for the marginalisation of
the cluster parameters (topics allocations) and mixing proportions.
This allowed \citet{sung2008latent} to derive a general form for
latent variable models, though our formulation is general to any
conjugate exponential graph.

\section{ Riemannian Gradient Based Optimisation}
\label{sec:riemann}
%The usual optimisation technique for VB models, sometimes called VBEM, is a co-ordinate ascent procedure which iterates between analytical optimisation of each variational factors using equation \eqref{eq:updates}. Previously, \citet{honkela2010approximate} proposed an approximate Riemannian optimisation scheme which showed some improvement over the VBEM algorithm. The basis of the idea is to perform conjugate gradient ascent in the space of a Riemannian manifold, whose co-ordinates are given by the parameters of the approximating distribution $q(\bZ)$ and whose curvature is given by the Fisher information. 
\citet{sato2001online} and \citet{hoffman2012stochastic} showed that the VBEM procedure performs
gradient ascent in the space of the natural parameters. % Recently,
% \citet{honkela2010approximate} proposed the use of a natural gradient
% procedure which connected the natural gradients to a convenient
% parameterisation of the problem by use of the Fisher information
% matrix. This follows from the information geometry
% \citep{amari2007methods} of the approximating distribution $q(\bZ)$,
% which forms a Riemannian manifold. The curvature of the manifold is
% dependent on the parameterisation of the problem, and is given by the
% Fisher information matrix.
Using the KLC bound to collapse the problem, gradient methods seem a
natural choice for optimisation, since there are fewer parameters to
deal with, and we have shown that computation of the gradients is
straightforward (the variational update equations contain the model
gradients).  It turns out that the KLC bound is particularly amenable
to {\em Riemannian} or {\em natural gradient} methods, because the
information geometry of the exponential family distrubution(s), over
which we are optimising, leads to a simple expression for the natural
gradient.  Previous investigations of natural gradients for
variational Bayes \citep{honkela2010approximate, kuusela2009gradient}
required the inversion of the Fisher information at every step (ours does not), and
also used VBEM steps for {\em some} parameters and Riemannian
optimisation for other variables. The collapsed nature of the KLC
bound means that these VBEM steps are unnecessary: the bound can be
computed by parameterizing the distribution of only one set of
variables ($q(\bZ)$) whilst the implicit distribution of the other
variables is given in terms of the first distribution and the data by
equation \eqref{eq:qstar_relation}. 

We optimize the lower bound $\Lkl$ with respect to the
parameters of the approximating distribution of the non-collapsed
variables. We showed in section \ref{sec:mvb} that the gradient of the
KLC bound is given by the gradient of the standard MF variational
bound, after an update of the {\em collapsed} variables. It is clear
from their definition that the same is true of the natural gradients.

%VB update equations, however %In general, the parameters of the approximating distribution are coupled, and the effect of making changes to $\rho$ on the bound is dependent on the current value of $\rho$. For example, adding a positive quantity to $\rho_{n1}$ is equivalent to subtracting that quantity from $\rho_{n(2\ldots K)}$. Alternatively, were the approximation a Gaussian distribution, the effect of changes to the mean vector would depend on the current value of the covariance. 

\subsection{Variable Transformations}

We can compute the natural gradient of our collapsed bound by
considering the update equations of the non-collapsed problem as
described above. However, if we wish to make use of more powerful
optimisation methods like conjugate gradient ascent, it is helpful
to re-parameterize the natural parameters in an unconstrained
fashion. The natural gradient is given by \citep{amari2007methods}:
\begin{equation}
\widetilde {\bf g}(\btheta) = G(\btheta)^{-1}\fracpartial{\Lkl}{\btheta}
\end{equation}
where $G(\btheta)$ is the Fisher information matrix whose $i$,$j^\text{th}$ element is given by 
\begin{equation}
G(\btheta)_{[i,j]} = - \E {q(\bX\given \btheta)} {\fracpartialij{\ln q(\bX\given\btheta)}{\btheta}}.
%G(\btheta) = - \E {q(\bX\given \btheta)} {\nabla^2_\btheta\ln q(\bX\given\btheta)}
\end{equation}
For exponential family distributions, this reduces to
$\nabla^2_\btheta\psi(\btheta)$, where $\psi$ is the log-normaliser.
Further, for exponential family distributions, the Fisher information
in the canonical parameters $(\btheta)$ and that in the {\em
  expectation} parameters $(\bEta)$ are reciprocal, and we also have $
G(\btheta) = \partial \bEta/ \partial \btheta $.  This means that the
natural gradient in $\btheta$ is given by
\begin{equation}
\widetilde {\bf g}(\btheta) = G(\btheta)^{-1} \fracpartial {\bEta}{\btheta} \fracpartial{\Lkl}{\bEta} = \fracpartial{\Lkl}{\bEta}\,\,\,\text{and}\,\,\,
\widetilde {\bf g}(\bEta) = \fracpartial{\Lkl}{\btheta}.  
\label{eq:natgrad_transform}
\end{equation}
The gradient in one set of parameters provides the natural gradient in the other.   Thus when our approximating distribution $q$ is exponential family, we can compute the natural gradient \emph{without} the expensive matrix inverse. 

\subsection{Steepest Ascent is Coordinate Ascent}
%\neil{This is unclear, we need to summarize the work of Sato and describe how it applies in our case. This is an important piece of the puzzle in describing why the KLC bound is so useful: we need to finish the story by saying VBEM is just steepest ascent on the KLC bound (as well as being coordinate ascent. So basically even if you thought you \emph{weren't} doing a gradient based method you were!!!}
%\neil{eq:natgrad\_simple and eq:natgrad\_general no longer exist.} JH:done
\citet{sato2001online} showed that the VBEM algorithm was a gradient based
algorithm. In fact, VBEM consists of taking {\em unit} steps in the direction
of the natural gradient of the canonical parameters. From equation
\eqref{eq:MVB_grad} and the work of \citet{sato2001online}, we see that
the gradient of the KLC bound can be obtained by considering the standard
mean-field update for the non-collapsed parameter $\bZ$. We confirm these
relationships for the models studied in the next section in the supplementary
material. 

Having confirmed that the VB-E step is equivalent to steepest-gradient
ascent we now explore whether the procedure could be improved by the
use of conjugate gradients.

\subsection{Conjugate Gradient Optimization}

One idea for solving some of the problems associated with steepest
ascent is to ensure each gradient step is conjugate (geometrically) to
the previous. \citet{honkela2010approximate} applied conjugate
gradients to the standard mean field bound, we expect much faster
convergence for the KLC bound due to its differing curvature. Since
VBEM uses a step length of 1 to optimize,\footnote{We empirically
  evaluated a line-search procedure, but found that in most cases that
  Wolfe-Powell conditions were met after a single step of unit
  length.} we also used this step length in conjugate gradients.
In the natural conjugate gradient method, the search direction at the
$i^\text{th}$ iteration is given by $\bs_i = -\widetilde \bg_i + \beta \bs_{i-1}$.
%\begin{equation}
  %\bs_i = -\widetilde \bg_i + \beta \bs_{i-1}. 
%\end{equation}
Empirically the Fletcher-Reeves method for estimating $\beta$ worked well for us:
\begin{equation}
	\beta_{FR} = \frac{\<\widetilde \bg_i, \widetilde \bg_i\>_i}{\<\widetilde \bg_{i-1}, \widetilde \bg_{i-1}\>_{i-1}}\\
\end{equation}
%where $\beta$ is found by either the Polack-Ribi\`ere, Fletch-Reeves or Hestenes-Stiefel methods:
%\begin{equation}
	%\begin{split}
		%\beta_{PR} = \frac{\<\widetilde \bg_i, \widetilde \bg_i-\widetilde \bg_{i-1}\>_i}{\<\widetilde \bg_{i-1}, \widetilde \bg_{i-1}\>_{i-1}}\\
		%\beta_{FR} = \frac{\<\widetilde \bg_i, \widetilde \bg_i\>_i}{\<\widetilde \bg_{i-1}, \widetilde \bg_{i-1}\>_{i-1}}\\
		%\beta_{HS} = \frac{\<\widetilde \bg_i, \widetilde \bg_i-\widetilde \bg_{i-1}\>_i}{\<\widetilde \bg_{i-1}, \widetilde \bg_i-\widetilde \bg_{i-1}\>_{i-1}}
	%\end{split}
%\end{equation}
where $\<\cdot, \cdot\>_i$ denotes the inner product in Riemannian
geometry, which is given by $\widetilde \bg^\top G(\rho) \widetilde
\bg$. We note from \citet{kuusela2009gradient} that this can be
simplified since $\widetilde \bg^\top G\widetilde \bg = \widetilde
\bg^\top GG^{-1}\bg = \widetilde \bg^\top \bg$, and other conjugate methods, defined in the supplementary material, can be applied similarly. 
%Since the natural gradient can be computed efficiently through
%equation \eqref{eq:natgrad_transform}, and the expression follows that for a VBEM update, the additional computational
%overhead involved in using the conjugate gradient method over the VBEM
%method is very small.

\section{Experiments}

For empirical investigation of the potential speed ups we selected a range of
probabilistic models. We provide derivations of the bound and fuller
explanations of the models in the supplementary material.  In each experiment,
the algorithm was considered to have converged when the change in the bound or
the Riemannian gradient reached below $10^{-6}$. Comparisons between
optimisation procedures always used the same initial conditions (or set of
initial conditions) for each method.  First we recreate the mixture of Gaussians example described by \citet{honkela2010approximate}.

\subsection{Mixtures of Gaussians}

For a mixture of Gaussians, using the d-separation rule, we select for
$\bX$ the cluster allocation (latent) variables. These are parameterised
through the softmax function for unconstrained optimisation.  Our model
includes a fully Bayesian treatment of the cluster parameters and the
mixing proportions, whose approximate posterior distributions appear as
\eqref{eq:qstar_relation}.  Full details of the algorithm derivation are
given in the supplementary material. A neat feature is that we can make
use of the discussion above to derive an
expression for the natural gradient without a matrix inverse. 

In \citet{honkela2010approximate} data are drawn from a mixture of
five two-dimensional Gaussians with equal weights, each with unit
spherical covariance. The centers of the components are at $(0,0)$ and
$(\pm R, \pm R)$. $R$ is varied from $1$ (almost completely
overlapping) to $5$ (completely separate). The model is initialised with eight components with an uninformative prior over the mixing proportions: the optimisation procedure is left to select an appropriate number of components. 

\citet{sung2008latent} reported that their collapsed method led to
improved convergence over VBEM. Since our objective is identical,
though our optimisation procedure different, we devised a metric for
measuring the efficacy of our algorithms which also accounts for their
propensity to fall into local minima. Using many randomised restarts,
we measured the average number of iterations taken to reach the
\emph{best-known optimum}. If the algorithm converged at a lesser
optimum, those iterations were included in the denomiator, but we
didn't increment the numerator when computing the average. We compared
three different conjugate gradient approaches and standard VBEM (which
is also steepest ascent on the KLC bound) using 500 restarts. 
%For each
%restart we used the same initial conditions for all algorithms.

Table \ref{tab:mog} shows the number of iterations required (on
average) to come within 10 nats of the best known solution for three
different conjugate-gradient methods and VBEM. VBEM sometimes failed
to find the optimum in any of the 500 restarts. Even relaxing the
stringency of our selection to 100 nats, the VBEM method was always at
least twice as slow as the best conjugate method.

\begin{table}
\caption{\small Iterations to convergence for the mixture of Gaussians problem, with varying
overlap (R). This table reports the average number of iterations taken to reach
(within 10 nats of) the best known solution. For the more difficult scenarios (with
more overlap in the clusters) the VBEM method failed to reach the optimum solution within 500
restarts}
\label{tab:mog}
	\vspace{3mm}
\begin{center}
\begin{tabular}{c|c|c|c|c|c}
CG. method& $R=1$ & $R=2$ & $R=3$ & $R=4$ & $R=5$ \\
\hline
Polack-Ribi\'ere & $3,100.37$ & $15,698.57$ & $5,767.12$ & $1,613.09$ & $3,046.25$\\
Hestenes-Stiefel & $1,371.55$ & $5,501.25$ & $5,922.4$ & $\mathbf{358.03}$ & $\mathbf{172.39}$\\
Fletcher-Reeves & $\mathbf{416.18}$ & $\mathbf{1,161.35}$ & $\mathbf{5,091.0}$ & $792.10$ & $494.24$\\
VBEM & $\infty$ & $\infty$ & $\infty$ & $992.07$ & $429.57$\\
\hline
\end{tabular}
\end{center}
\end{table}

\subsection{Topic Models}

Latent Dirichlet allocation (LDA) \citep{blei2003latent} is a popular approach for extracting topics from documents.
%is a model for
% interrogating corpora of documents, using the idea of a latent set of topics
% to explain the occurrences of words across the documents. A derivation of the
% KLC bound and its gradient are given in the supplementary material.  To test
% the effectiveness of our algorithm,
To demonstrate the KLC bound we applied it to 200 papers from the 2011
NIPS conference. The PDFs were preprocessed with \texttt{pdftotext},
removing non-alphabetical characters and coarsely filtering words by
popularity to form a vocabulary size of 2000.\footnote{Some extracted
  topics are presented in the supplementary material.} We selected the
latent topic-assignment variables for parameterisation, collapsing the
topics and the document proportions. Conjugate gradient
optimization was compared to the standard VBEM approach.

We used twelve random initializations, starting each algorithm from
each initial condition.  Topic and document distributions where
treated with fixed, uninformative priors. On average, the
Hestenes-Steifel algorithm was almost ten times as fast as standard
VB, as shown in Table \ref{tab:lda}, whilst the final bound varied
little between approaches.

\begin{table}
  \caption{\small Time and iterations taken to run LDA on the NIPS 2011 corpus, $\pm$ one standard deviation, for two conjugate methods and VBEM. The Fletcher-Reeves conjugate algorithm is almost ten times as fast as VBEM. The value of the bound at the optimum was largely the same: deviations are likely just due to the choice of initialisations, of which we used 12. }
  \label{tab:lda}
	\vspace{3mm}
\begin{center}
\begin{tabular}{c|ccc}
Method & Time (minutes) & Iterations & Bound\\
\hline
Hestenes-Steifel & $ 56.4 \pm 18.5 $ & $ 644.3 \pm 214.5 $ & $ -1,998,780 \pm 201 $\\
Fletcher-Reeves & $ \mathbf{38.5 \pm 8.7} $ & $ \mathbf{447.8 \pm 100.5} $ & $ -1,998,743 \pm 194 $\\
VBEM & $370 \pm 105 $ & $4,459 \pm 1,296$ & $-1,998,732 \pm 241$\\
\hline
\end{tabular}
\end{center}
\end{table}

\subsection{RNA-seq alignment} 

An emerging problem in computational biology is inference of transcript structure and
expression levels using next-generation sequencing technology (RNA-Seq).  Several
models have been proposed. The BitSeq method \citep{glaus2012} is
based on a probabilistic model and uses Gibbs sampling for approximate
inference. The sampler can suffer from particularly slow convergence
due to the large size of the problem, which has six million latent
variables for the data considered here. We implemented a variational
version of their model and optimised it using VBEM and our collapsed
Riemannian method.  We applied the model to data described in
\citet{xu2010a}, a study of human microRNA. The model was initialised
using four random initial conditions, and optimised using standard
VBEM and the conjugate gradient versions of the algorithm. The
Polack-Ribi\'ere conjugate method performed very poorly for this
problem, often giving negative conjugation: we omit it here. The
solutions found for the other algorithms were all fairly close, with
bounds coming within 60 nats. The VBEM method was dramatically
outperformed by the Fletcher-Reeves and Hestenes-Steifel methods: it
took $4600\pm 20$ iterations to converge, whilst the conjugate methods
took only $268 \pm 4$ and $265 \pm 1$ iterations to converge. At about
8 seconds per iteration, our collapsed Riemannian method requires
around forty minutes, whilst VBEM takes almost eleven hours. All the
variational approaches represent an improvement over a Gibbs sampler,
which takes approximately one week to run for this data
\citep{glaus2012}.

\section{Discussion}

Under very general conditions (conjugate exponential family) we have
shown the equivalence of collapsed variational bounds and marginalized
variational bounds using the KL corrected perspective of
\citet{king2006fast}. We have provided a succinct derivation of these
bounds, unifying several strands of work and laying the foundations for
much wider application of this approach.

When the collapsed variables are updated in the standard MF bound the
KLC bound is identical to the MF bound in value and
gradient. \citet{sato2001online} has shown that coordinate ascent of
the MF bound (as proscribed by VBEM updates) is equivalent to steepest
ascent of the MF bound using natural gradients. This implies that
standard variational inference is also performing steepest ascent on
the KLC bound. This equivalence between natural gradients and the VBEM
update equations means our method is quickly implementable for any
model where the mean field update equations have been computed. It is
only necessary to determine which variables to collapse using a
d-separation test. Importantly this implies our approach can readily
be incorporated in automated inference engines such as that provided
by infer.net \citep{minka2010infer}.
% In applying our Riemannian conjugate gradient optimisation routine
% (inspired by that of \citet{honkela2010approximate}) to the KLC bound
% we made use of some properties of the information geometry 
% exponential families and saw dramatically improved convergence
% performance across a range of different probabilistic models.
We'd like to emphasise the ease with which the method can be applied:
we have provided derivations of equivalencies of the bounds and
gradients which should enable collapsed conjugate optimisation of
{\em any} existing mean field algorithm, with minimal changes to the
software. Indeed our own implementations (see supplementary material)
use just a few lines of code to switch between the VBEM and conjugate
methods. 

The improved performance arises from the curvature of the KLC
bound. We have shown that it is always less negative than that of the
original variational bound allowing much larger steps in the
variational parameters as \citet{king2006fast} suggested. This also
provides a gateway to second-order optimisation, which
%when combined with insights from information geometry as we have here,
could prove even faster.

We provided empirical evidence of the performance increases that are
possible using our method in three models.  In a thorough exploration
of the convergence properties of a mixture of Gaussians model, we
concluded that a conjugate Riemannian algorithm can find solutions
that are not found with standard VBEM. In a large LDA model, we found
that performance can be improved many times over that of the VBEM
method. In the BitSeq model for differential expression of genes
transcripts we showed that very large improvements in performance are
possible for models with huge numbers of latent variables.

%We provided empirical evidence that improved convergence speed is
%possible by applying our ideas in a Bayesian MoG model, making use of
%cojugate gradients for optimization. In this case the non-collapsed
%variable is represented by a multinomial distribution, and we provided
%a simple connection between the natural gradients and the softmax
%parameterisation. This approach could easily be extended to related
%models such as LDA.

\section*{Acknowledgements}
The authors would like to thank Michalis Titsias for helpful commentary on a previous draft and Peter Glaus for help with a C++ implementation of the RNAseq alignment algorithm. This work was funded by EU FP7-KBBE Project Ref 289434 and BBSRC grant number BB/1004769/1. 

\clearpage
\bibliographystyle{abbrvnat}
\bibliography{main}

\clearpage

\begin{center}
{\Large Supplementary Material for:\\ Fast Variational Inference \\in the Conjugate Exponential Family}
\end{center}
This supplementary material accompanies the NIPS paper on Fast
Variational Inference in the Conjugate Exponential Family. Its purpose
is to provide details of how our very general framework applies in the
case of the specific models described in the paper.  First we briefly
mention the form of the three conjugate gradient algorithms we used in
optimization.

\section{Conjugate gradient algorithms}
There are several different methods for approximating the parameter $\beta$ in the conjugate gradient algorithm. We used the Polack-Ribi\`ere, Fletch-Reeves or Hestenes-Stiefel methods:
\begin{equation}
	\begin{split}
		\beta_{PR} = \frac{\<\widetilde \bg_i, \widetilde \bg_i-\widetilde \bg_{i-1}\>_i}{\<\widetilde \bg_{i-1}, \widetilde \bg_{i-1}\>_{i-1}}\\
		\beta_{FR} = \frac{\<\widetilde \bg_i, \widetilde \bg_i\>_i}{\<\widetilde \bg_{i-1}, \widetilde \bg_{i-1}\>_{i-1}}\\
		\beta_{HS} = \frac{\<\widetilde \bg_i, \widetilde \bg_i-\widetilde \bg_{i-1}\>_i}{\<\widetilde \bg_{i-1}, \widetilde \bg_i-\widetilde \bg_{i-1}\>_{i-1}}
	\end{split}
\end{equation}
where $\<\cdot, \cdot\>_i$ denotes the inner product in Riemannian
geometry, which is given by $\widetilde \bg^\top G(\rho) \widetilde
\bg$

\section{Mixture of Gaussians}
\label{sec:mog}
A MoG model is defined as follows.  We have a set of
$N$ $D$-dimensional vectors $\bY = \{\by_n\}_{n=1}^N$.  The likelihood
is
\begin{equation}
	p(\bY|\bEta, \bL) = \prod_{k=1}^K\prod_{n=1}^N \mathcal N(\by_n|\bmu_k, \bLambda_k^{-1})^{\ell_{nk}}
\end{equation}
where $\bL$ is a collection of binary latent variables indicating
cluster membership, $\bL = \{\{\bell_{nk}\}_{n=1}^N\}_{k=1}^K$ and $\bEta$
is a collection of cluster parameters, $\bEta = \{\bmu_k,
\bLambda_k\}_{k=1}^K$

The prior over $\bL$ is given by a multinomial distribution with components $\bpi$, which in turn have a Dirichlet prior with uniform concentrations for simplicity:
\begin{equation}
	p(\bL|\bpi) = \prod_{k=1}^K\prod_{n=1}^N \pi_k^{\ell_{nk}},\hspace{0.43cm}
	p(\bpi) = R_{D}(\balpha)\prod_{k=1}^K \pi_k^{\alpha-1}
\end{equation}
%\begin{equation}
	%p(\bL|\bpi) = \prod_{k=1}^K\prod_{n=1}^N \pi_k^{\bell_{nk}}
%\end{equation}
%\begin{equation}
	%p(\bpi) = R_{Di}(\balpha)\prod_{k=1}^K \pi_k^{\alpha-1}
%\end{equation}
with $\balpha$ representing a $K$ dimensional vector with elements $\alpha$, and $R_{D}$ being the normalising constant for the Dirichlet distribution, $R_{D}(\balpha) = \Gamma(K \alpha)\Gamma(\alpha)^{-K}$.  

Finally we choose a conjugate Gaussian-Wishart prior for the cluster parameters 
%\begin{equation}
	%p(\bEta) = \prod_{k=1}^K\mathcal N(\bmu_k|\bm_0, (\kappa_0\bLambda_k)^{-1}) \mathcal W(\bLambda_k|\bS_0, \nu_0)
%\end{equation}
which can be written 
\begin{equation}
\begin{split}
	\ln p(\bmu_k, \bLambda_k) = \ln R_{GW}(\bS_0,\nu_0,\kappa_0) +   \frac{\nu_0-D}{2}\ln |\bLambda_k|\\
	-\frac{1}{2}\text{tr}\left(\bLambda_k\left( \kappa_0\bmu_k\bmu_k^\top + \kappa_0\bm_0\bm_0^\top - 2\kappa_0\bm_0\bmu_k^\top + \bS_0\right)\right)
\end{split}
\end{equation}
where $R_{GW}$ is the normalising constant, and is given by 
\[ R_{GW}(\bS, \nu,\kappa) = |\bS|^\frac{\nu}{2} 2^{-\frac{(\nu+1)D}{2}}  \pi^{-\frac{D(D+1)}{4}} \kappa^\frac{D}{2} (\prod_{d=1}^D\Gamma( (\nu+1-d)/2))^{-1}.
\]  
%$ \ln R_{GW}(\bS, \nu,\kappa) = \frac{\nu}{2}\ln|\bS| -\frac{(\nu+1)D}{2}\ln2 - \frac{D(D+1)}{4}\ln \pi - \sum_{d=1}^D\ln\Gamma( (\nu+1-d)/2) + \frac{D}{2}\ln \kappa$.  

\subsection{Applying the KLC bound}
\begin{figure}
	\centering
	\includegraphics[width=0.5\textwidth]{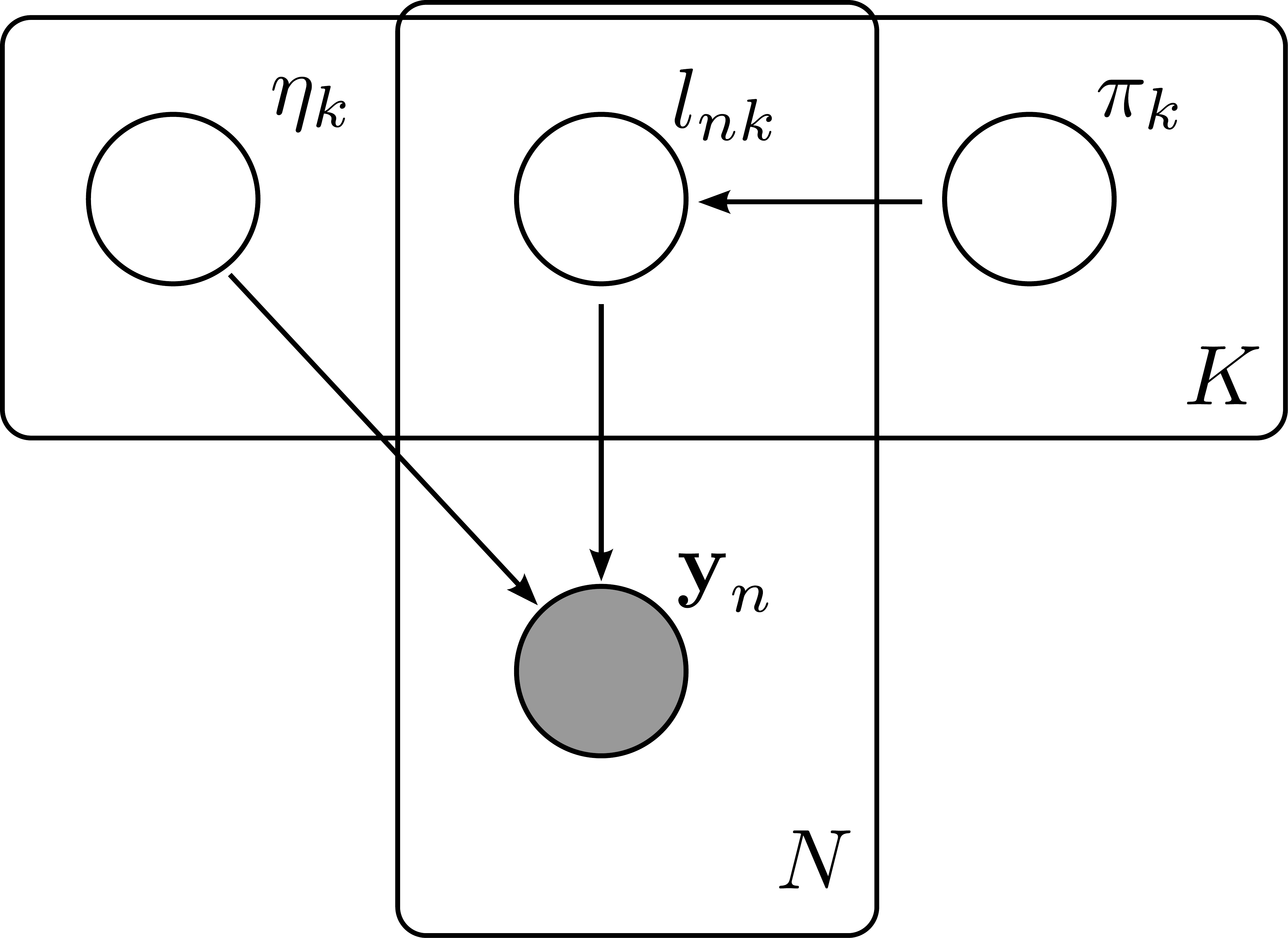}
	\caption{\label{fig:mog_graphical} A graphical model representation of the MoG model. A d-separation test quickly shows that it is possible to marginalise $\bpi$ and $\bEta$ given a variational parameterisation of $\bL$. }
\end{figure}

The first task in applying the KLC bound is to select which variables to parameterise and which to marginalise. From the graphical model representation of the MoG problem in Figure \ref{fig:mog_graphical}, we can see that we can select the latent variables $\bZ = \{\bL\}$ for parameterisation, whilst marginalising the mixing proportions and cluster parameters ($\bX = \{\bpi, \bEta\}$). We note that it is possible to select the variables the other way around: parameterising $\bpi$ and $\bEta$ and marginalising $\bL$, but parameterisation of the latent variables makes implementation a little simpler. 

We use a factorised multinomial distribution $q(\bL)$ to approximate the posterior for $p(\bL|\bY)$, parameterised using the softmax functions so
\begin{equation}
  q(\bL) = \prod_{n=1}^N\prod_{k=1}^K r_{nk}^{\bell_{nk}}\,\,,\,\,\,\,\,r_{nk} = \frac{e^{\rho_{nk}}}{\sum_{i=1}^K e^{\rho_{ni}}}.  
	\label{eq:softmax}
\end{equation}
We are now ready to apply the procedure described above to derive the KLC bound. First, 
\begin{equation}
	\ln p(\bY|\bEta, \bpi) \geq \int\! q(\bL) \{ \ln p(\bY|\bEta, \bL) + \ln p(\bL|\bpi)\}\,\text d\bL + H_L,
\end{equation}
where $H_L$ is the entropy of the distribution $q(\bL)$. We expand to give
\begin{equation}
	\begin{split}
		%\mathcal L_1 = -\frac{ND}{2}\ln(2\pi) + \frac{1}{2} \sum_{k=1}^K\bigg\{ \rhat_k\ln |\bLambda_k| \\
		%- \text{tr}\big(\bLambda_k \big( \rhat_k\bmu_k\bmu_k^\top + C_k -2\bmu_k\ybar_k\big)\big)\\
		%+ \rhat_k\ln\pi_k\bigg\} + H_L\\
\mathcal L_1 =  \tfrac{1}{2} \sum_{k=1}^K\Big\{- \text{tr}\big(\bLambda_k \big( \rhat_k\bmu_k\bmu_k^\top + C_k -2\bmu_k\ybar_k\big)\big)\\
+\rhat_k\ln\pi_k + \rhat_k\ln |\bLambda_k|  \Big\} + H_L -\tfrac{ND}{2}\ln(2\pi) 
	\end{split}
\end{equation}
where $\rhat_k = \sum_{n=1}^N r_{nk}$, $C_k = \sum_{n=1}^N r_{nk}
\by_n\by_n^\top$, and $\ybar_k = \sum_{n=1}^N r_{nk}\by_n$.  The
conjugacy between the intermediate bound $\mathcal L_1$ and the prior
now emerges, making the second integral in the KLC bound tractable.

After exponentiating this expression and multiplying by the prior,
$p(\bEta)p(\bpi)$, we find that the integrals with respect to both
$\bEta$ and $\bpi$ are tractable. This result means that the only
variational parameters needed are those of $q(\bL)$.  
%\magnus{below you say the
  %true variational parameters $r$. Do you mean $\rho$ or $r$ or does
  %it not matter?}.  
The integrals result in
\begin{equation}
\begin{split}
	\mathcal L_\text{KL} = -\frac{ND}{2}\ln(2\pi) + \ln R_{Di}(\balpha) -\ln R_{Di}(\balpha')\\ + K\ln R_{GW}(\bS_0,\nu_0,\kappa_0) - \sum_{k=1}^K\ln R_{GW}(\bS_k, \nu_k, \kappa_k) + H_L
\end{split}
\label{eq:bound_1}
\end{equation}
where we have defined 
%\begin{align}
  %\alpha_k &= \alpha + \rhat_k\label{eq:updates_1}\\
  %\kappa_k &= \kappa_0 + \rhat_k\\
  %\nu_k &= \nu_0 + \rhat_k\\
  %\bm_k &= (\kappa_0\bm_0 + \ybar_k)/\kappa_k\\
  %\bS_k &= S_0 + C_k + \kappa_0\bm_0\bm_0^\top -\kappa_k\bm_k\bm_k^\top\label{eq:updates_2}, 
%\end{align}
\begin{equation}
\begin{split}
\alpha_k &= \alpha + \rhat_k \hspace{2.5cm}   \kappa_k = \kappa_0 + \rhat_k\\
\bm_k &= (\kappa_0\bm_0 + \ybar_k)/\kappa_k \hspace{1cm}  \nu_k = \nu_0 + \rhat_k\\
\bS_k &= S_0 + C_k + \kappa_0\bm_0\bm_0^\top -\kappa_k\bm_k\bm_k^\top
\label{eq:M_steps}
\end{split}
\end{equation}
and $\balpha'$ represents a vector containing each $\alpha_k$.  Some simplification of \eqref{eq:bound_1} leads to
\begin{equation}
	\begin{split}
		\mathcal L_\text{KL} &= \sum_{k=1}^K\Big\{ \ln\Gamma(\alpha_k)-\tfrac{D}{2}\ln \kappa_k -\tfrac{\nu_k}{2}\ln |S_k| \\
	&+ \sum_{d=1}^D \ln \Gamma( (\nu_k+1-d)/2 ) \Big\}  + H_L +\text{const.}
	\label{eq:MVB_simple}
	\end{split}
\end{equation}
where {const.} contains terms independent of $\br$.%, and equates to $ -\frac{ND}{2}\ln\pi +  \frac{KD}{2}\ln\kappa_0 +\frac{K\nu_0}{2}\ln|S_0| +\ln\Gamma(K\alpha) -K\ln\Gamma(\alpha) + \ln\Gamma(K\alpha+N)- K\sum_{d=1}^D\ln\Gamma( (\nu +1-d)/2)$. 

Equations \eqref{eq:M_steps} are similar to the update equations for
the approximating distributions in the VBEM methodology \citep[see
e.g.][]{bishop2006pattern}. However, for our model they are simply
intermediate variables, representing combinations of the true
variational parameters $\br$, the data, and the model prior
parameters. When optimizing the model with respect to the variational
parameters, the dependency of these intermediate variables on $\br$ is
not ignored as it would be in MF variational approach.

The gradient of the MV bound \eqref{eq:MVB_simple} with respect to the parameters $\br$ is given by
\begin{equation}
	\begin{split}
		%\frac{\partial\mathcal L_{MV}}{\partial r_{nk}} =& -\frac{D}{2}\kappa_k^{-1} - \frac{1}{2}\ln|S_k| + \digamma(\alpha_k) -\ln r_{nk}-1\\ 
		\frac{\partial\mathcal L_\text{KL}}{\partial r_{nk}} =& -\tfrac{D}{2}\kappa_k^{-1} - \tfrac{1}{2}\ln|S_k| + \digamma(\alpha_k) -\ln r_{nk}\\ 
		&-\tfrac{\nu_k}{2}(y_n-\bm_k)^\top S_k^{-1}(y_n-\bm_k))\\
		&+\tfrac{1}{2}\sum_{d=1}^D\digamma( (\nu_k+1-d)/2)-1.
\end{split}
\label{eq:gradient}
\end{equation}
Taking a step in this direction (in the valiables $\gamma$) yields exactly the VB-E step associated with the mean-field bound. the gradient in $r$ is the natural gradient in $\gamma$ (see paper section 4.1).

\section{Latent Dirichlet Allocation}
Latent Dirichlet allocation is a popular topic model. See \citet{blei2003latent} for a thorough introduction. 

%Nomenclature\\
Suppose we have $D$ documents, $K$ topics and a vocabulary of size $V$. The $d^\text{th}$ document contains $N_d$ words $W_d = \{w_{dn}\}_{n=1}^{N_d}$, and each word is represented as a binary vector $w_{dn} \in \{0,1\}^V$.  Each word is associated with a latent variable $\bell_{dn}$, which assigns the word to a topic, thus $\bell_{dn} \in \{0,1\}^K$.  We'll use $W$ to represent the colletion of all words, $W = \{W_d\}_{d=1}^D$, and $\bL$ to represent the collection of all latent variables $\bL = \{\{\bell_{dn}\}_{n=1}^{N_d}\}_{d=1}^D$.  

Each document has an associated vector of topic proportions, $\theta_d \in [0,1]^K$, and each topic is represented by a vector of word proportions $\phi_k \in [0,1]^V$. We assume a symmetrical prior distribution over topics in each document $p(\theta_d) = \Dir (\theta_d|\alpha)$, and similarly for words within topics. $p(\phi_k) = \Dir(\phi_k|\beta)$. 

The LDA generative model states that for each word, first the associated topic is drawn from the topic proportions for the document, and then the word is drawn from the selected topic. 
\begin{equation}
\begin{split}
 p(\ell_{dn}|\theta_d) &= \prod_{k=1}^K \theta_{dk}^{\bell_{dnk}}\\
 p(w_{dn}|\bell_{dn},\phi) &= \prod_{k=1}^K\prod_{v=1}^V \phi_{kv}^{w_{dnv}\bell_{dnk}}\\
\end{split}
\end{equation}

\subsection{The collapsed bound}
\begin{figure}
	\centering
	\includegraphics[width=0.5\textwidth]{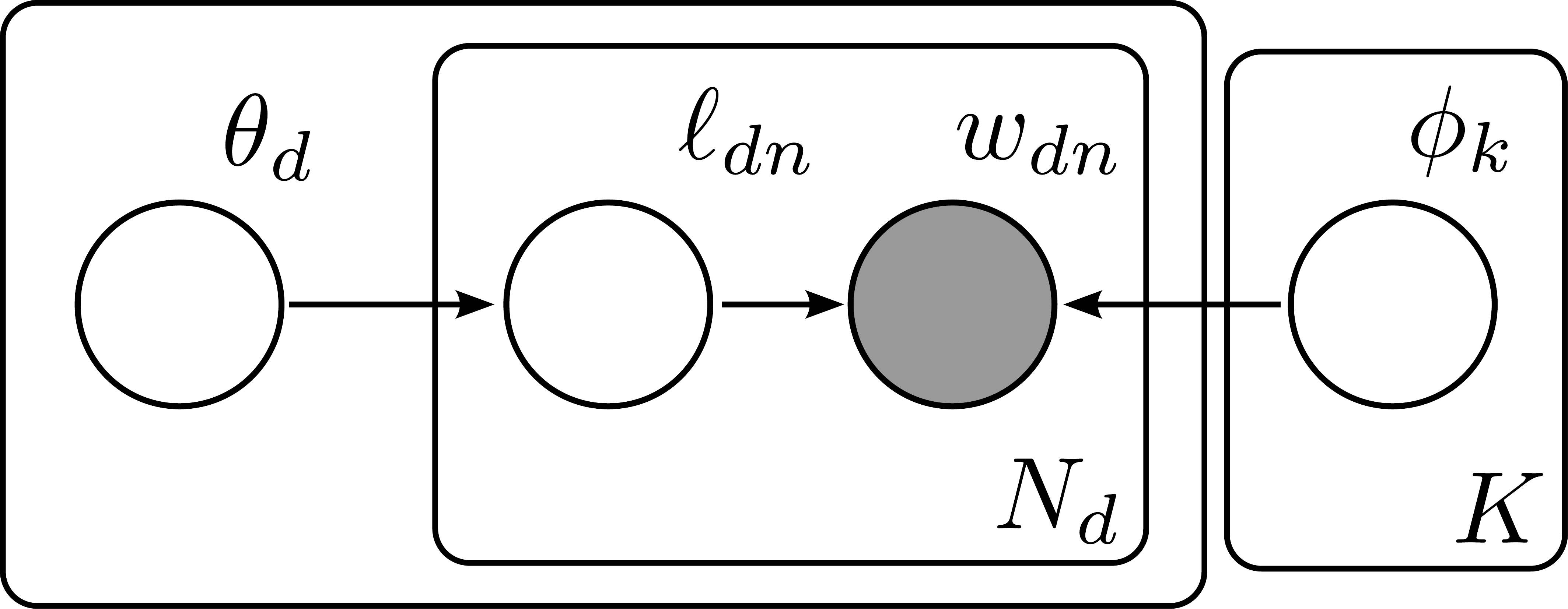}
	\caption{\label{fig:lda_graphical} A graphical model representation of Latent Dirichlet allocation. A d-separation test quickly shows that it is possible to marginalise $\theta$ and $\phi$ given a variational parameterisation of $\bL$. }
\end{figure}

To derive the colapsed bound, we use a similar d-separation test as for the mixture model to select the latent variables as the parameteriser (non-collapsed) nodes. See Figure \ref{fig:lda_graphical}. 

To proceed we assume a factorising multinomial posterior for $\bL$:
\begin{equation}
	q(\bL) = \prod_{d=1}^D \prod_{n=1}^{N_d}\prod_{k=1}^K r_{dnk}^{\bell_{dnk}}
\end{equation}
subject to the constraint $\sum_{k=1}^K\bell_{dnk} = 1$, which we enforce through a softmax reparameterisation
\begin{equation}
r_{dnk} = \frac{e^{\rho_{dnk}}}{\sum_{k'=1}^K e^{\rho_{dnk'}}}.
\end{equation}
We proceed by deriving the conditional bound 
\begin{equation}
	\ln p(W\given \theta, \phi) \geq \mathcal L_1 = \sum_{d=1}^D \sum_{n=1}^{N_d}\sum_{k=1}^K\sum_{v=1}^V (w_{dnv} r_{dnk})\ln \phi_{kv} + \sum_{d=1}^D \sum_{n=1}^{N_d}\sum_{k=1}^K(r_{dnk}\ln \theta_{dk}) + H[q(\bL)].  
\end{equation}

To marginalise the variables $\theta, \phi$, we exponentiate this bound and take the expectation under the priors. This results in 
\begin{equation}
\begin{split}
	p(W)\geq\int\!\exp\{\mathcal L_1 \}p(\theta)p(\phi)\d \theta \d \phi =& \int\!\prod_{k=1}^K\prod_{v=1}^V \phi_{kv} ^{\sum_{d=1}^d \sum_{n=1}^{n_d}(w_{dnv} r_{dnk})} \prod_{d=1}^D \prod_{k=1}^K\theta_{dk}^{(\sum_{n=1}^{N_d}r_{dnk})} \\
&\prod_{k=1}^KR_{Di}(\beta) \prod_{v=1}^V\phi_{kv}^{\beta-1}\\
&\prod_{d=1}^DR_{Di}(\alpha) \prod_{k=1}^K\theta_{dk}^{\alpha-1}\d \theta \d \phi \\
&\exp\{H[q(\bL)]\}.  
\end{split}
\end{equation}

Careful inspection of the above reveals that the two integrals separate as expected, and result in the  normalizers for each of the independent Dirichlet approximations. Taking the logarithm reults in
\begin{equation}
	\mathcal L_{\text{KL}} = D\ln R_{Di}(\alpha) - \sum_{d=1}^D \ln R_{Di}(\alpha_d') + K\ln R_{Di}(\beta) - \sum_{k=1}^K\ln R_{Di}(\beta_k') + H[q(\bL)]
\end{equation}
where we have defined $\alpha_{dk}' = \alpha + \sum_{n=1}^{N_d} r_{dnk}$ and $\beta_{kv}' = \beta + \sum_{d=1}^d \sum_{n=1}^{n_d}w_{dnv} r_{dnk}$.

\subsection{Topics found by LDA}

For completeness we show here some topics found by LDA on the NIPS conference data.
\begin{table}[!h]
\caption{some topics found using LDA on papers from the 2011 NIPS conference.  }
\label{tab:topics}
\vspace{3mm}
\begin{center}
\begin{tabular}{c|c|c|c|c|c}
neural & training & distribution & data & features & model \\
input &  feature &  gaussian & points & image & models \\
neurons &  classification & inference & point & object & variables \\
network &  class & process & clustering & images & parameters \\
fig &  tree & prior & distance & objects & structure \\
estimate &  prediction & sampling & dataset & scene & variable \\
neuron &  label & likelihood & similarity & recognition & markov \\
visual &  accuracy & posterior & cluster & reference & observed \\
nonlinear &  labels & distributions & manifold & detection & graphical \\
linear &  classifier & bayesian & spectral & part &hidden\\
\end{tabular}
\end{center}
\end{table}

\section{BitSeq Model}

The generative model for an RNA-seq assay is as follows. We assume that the experiment consists of a pile of RNA fragments, where the abundance of fragments from transcript $T_m$ in the assay is $\theta_m$. The sequencer then selects a fragment at random from the pile, such that the probability of picking a fragment corresponding to transcript $T_m$ is $\theta_m$. Introducing a convenient membership vector $\bell_n$ for each read,  we can write 
\begin{equation}
	p(\bL|\btheta) = \prd n \prd m \theta_m^{\bell_{nm}}
\end{equation}
where $\ell_{nm} \in \{0,1\}$ is a binary variable which indicates whether the $n$th fragment came from the $m$th transcript ($\ell_{nm}=1$) and is subject to $\sumover m \ell_{nm} = 1$. We use $\bL$ to represent the collection of all alignment variables. 

Both $\theta$ and $\bL$ are variables to be inferred, with $\btheta$ the main object of interest.

Writing the collection of all reads as $\bR = \{\br_n\}_{n=1}^N$, the likelihood of a set of alignments $\bL$ is
\begin{equation}
	p(\bR|\bT,\bL) = \prd n p(\br_n|T_m)^{\ell{nm}}
\end{equation}
where $T_m$ represents the $m$th transcript, $\bT$ represents the transcriptome.  

The values of $p(r_n|T_m)$ can be computed before performing inference in $\btheta$ since we are assuming a known transcriptome. We compute these values based on the quality of alignment of the read $\br_n$ to the transcript $T_m$, using a model which can correct for sequence specific or fragmentation biases. The method is described in detatil in \citet{glaus2012}.

%\paragraph{Prior over $\btheta$\\}
We specify a conjugate Dirichlet prior over the vector $\btheta$. 
\begin{equation}
p(\btheta) = \frac{\Gamma(\hao)}{\prd m \Gamma(\amo)} \prd m \theta_m ^{\amo-1}
\end{equation}
with $\hao = \sumover m \amo$. $\amo$ represents our prior belief in the values of $\theta_m$, and we use a relatively uninformative but proper prior $\amo=1\,\forall m=1\ldots M$. A priori, we assume that the concentrations are all equal, but with large uncertainty.  

\subsection{The collapsed bound}
\begin{figure}
	\centering
	\includegraphics[width=0.5\textwidth]{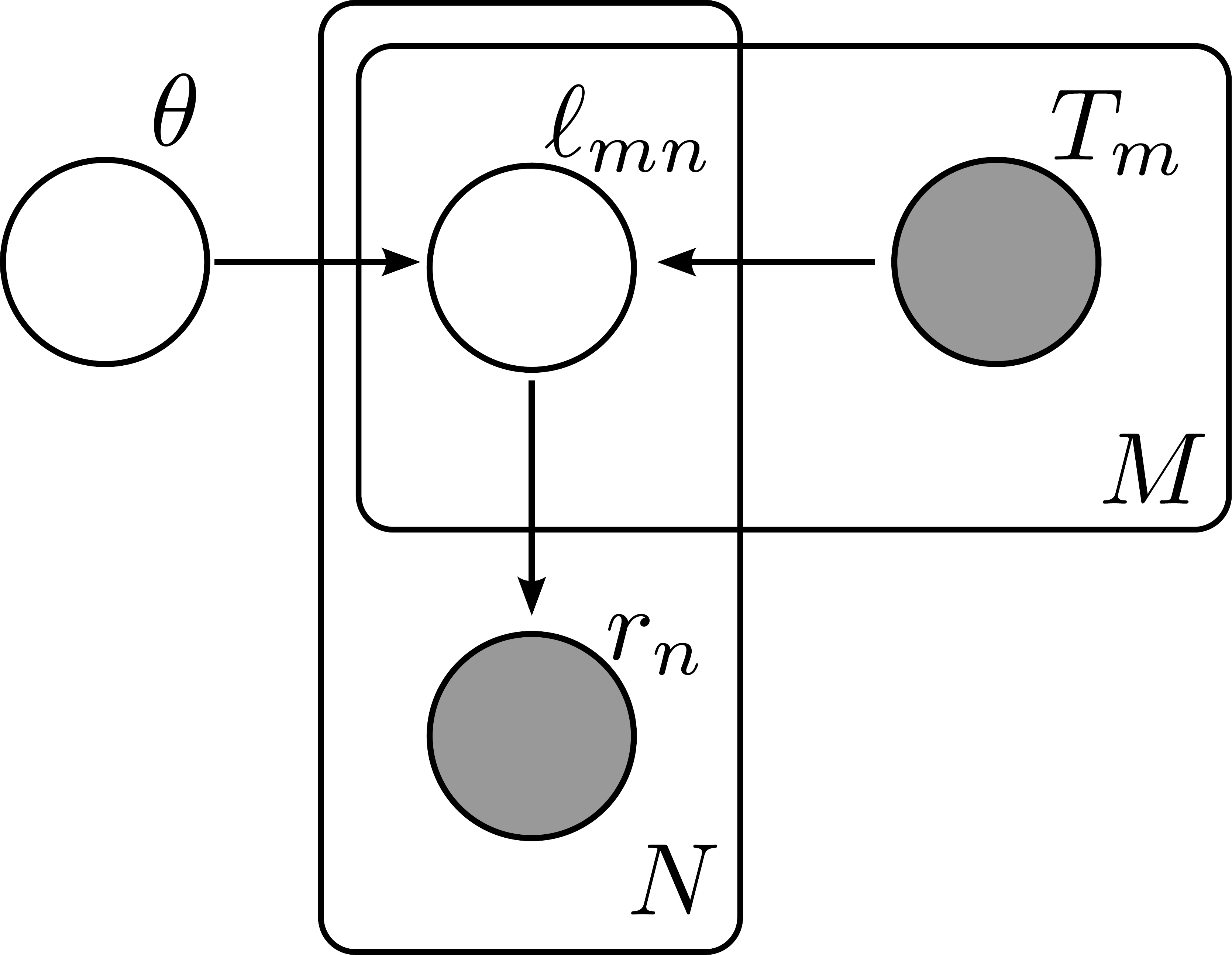}
	\caption{\label{fig:bitseq_graphical} A graphical model representation of the BitSeq model.  A d-separation test quickly shows that it is possible to marginalise $\theta$ given a variational parameterisation of $\bL$. }
\end{figure}

Figure \ref{fig:bitseq_graphical} shows a graphical representation of the BitSeq model. It's clear that parameterisation of the latent variables will allow us to collapse $\theta$, or vica-versa. Selecting again the latent variables for parameterisation, $\bX = \{\bL\}$, $\bZ = \{\theta\}$, we first find the conditional bound as usual by:
\begin{equation}
\begin{split}
\ln p(\bR \given \bT, \btheta) &= \ln \int\! p(\bR\given \bL, \bT) p(\bL\given\btheta) \d \bL\\
&\geq \E {q(\bL)} {\ln p(\bR\given \bL, \bT) + \ln p(\bL\given\btheta) - \ln q(\bL)}\\
&\geq \sumover n \sumover m \ell_{nm} \big(\ln p(\br_n\given T_m) + \ln \theta_m - \ln \ell_{nm}\big)\\
&\geq \mathcal L_1
\end{split}
\end{equation}

It's clear that this bound is conjugate to the prior for $\theta$, so we can marginalise:
\begin{equation}
\begin{split}
	\ln p(\bR\given \bT) \geq \mathcal L_\text{KL} = \sumover n \sumover m \ell_{nm} \big(\ln p(\br_n\given T_m) - \ln\ell_{nm} \big) + \ln \Gamma(\hao) -\ln \Gamma(\hao + N) \\-\sumover m \Big(\ln \Gamma(\amo)-\ln \Gamma(\amo + \hat\ell_m)\Big)
\end{split}
\label{eq:bound}
\end{equation}
where $\hat\ell_m = \sumover n \ell_n$ and we also have that the approximate posterior distribution for $\btheta$ is a Dirichlet distribution with parameters $\amo + \hat\ell_m$.

\end{document}